\pdfoutput=1

\documentclass[11pt]{article}

\usepackage[final]{acl}

\usepackage{times}
\usepackage{latexsym}

\usepackage[T1]{fontenc}

\usepackage[utf8]{inputenc}

\usepackage{microtype}

\usepackage{inconsolata}

\usepackage{enumitem}
\usepackage{booktabs}
\usepackage{graphicx}

\usepackage{subcaption}

\newcommand\wav{Wav2vec2}
\newcommand\bert{CamemBERT}   
\newcommand\wavbert{Wav2vec2 + CamemBERT}

\newcommand\hops{\texttt{hops}}
\newcommand\deplabel{\textit{dep2label}}

\newcommand\AUDIO{\textbf{\textsc{audio}}}
\newcommand\PIPELINE{\textbf{\textsc{pipeline}}}
\newcommand\ORACLE{\textbf{\textsc{oracle}}}
\newcommand\TEXT{\textbf{\textsc{text}}}
\newcommand{\SEQ}{\textsc{seq2label}}
\newcommand{\GRAPH}{\textsc{graph}}
\newcommand\HOPS{\textsc{hops}}

\title{Growing Trees on Sounds: Assessing Strategies \\ for End-to-End Dependency Parsing of Speech}

\author{Adrien Pupier, Maximin Coavoux, Jérôme Goulian, Benjamin Lecouteux \\
   Univ. Grenoble Alpes, CNRS, Grenoble INP, LIG, 38000 Grenoble, France\\
  \texttt{first.last@univ-grenoble-alpes.fr}
}

\begin{document}
\maketitle
\begin{abstract}
Direct dependency parsing of the speech signal --as opposed to parsing speech transcriptions-- has recently been proposed as a task \citep{pupier22_interspeech}, as a way of incorporating prosodic information in the parsing system and bypassing the limitations of a pipeline approach that would consist of using first an Automatic Speech Recognition (ASR) system and then a syntactic parser.
In this article, we report on a set of experiments aiming at assessing the performance of two parsing paradigms (graph-based parsing and sequence labeling based parsing) on speech parsing.
We perform this evaluation on a large treebank of spoken French, featuring
realistic spontaneous conversations.
Our findings show that (i) the graph-based approach obtain better results across the board (ii) parsing directly from speech outperforms a pipeline approach, despite having 30\% fewer parameters.

\end{abstract}

\begin{figure}
    \centering
    \begin{subfigure}{0.4\textwidth}
        \includegraphics[width=\textwidth]{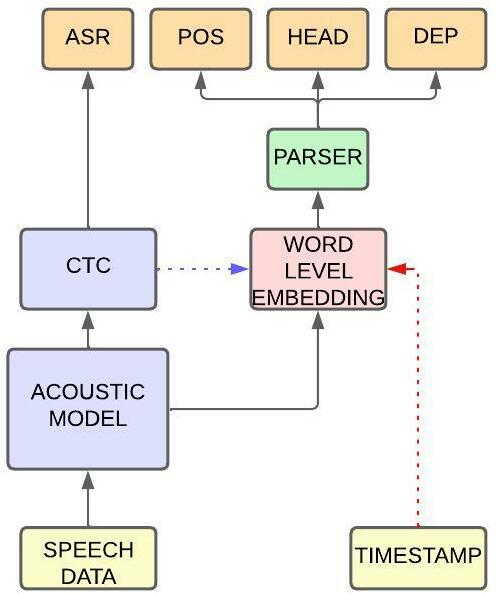}
        \caption{The two models based on audio features, blue arrow is \textcolor{blue}{\AUDIO{}}, red arrow is \textcolor{red}{\ORACLE{}}.}
        \label{fig:parser1}
    \end{subfigure}
    \hfill
    \begin{subfigure}{0.4\textwidth}
        \includegraphics[width=\textwidth]{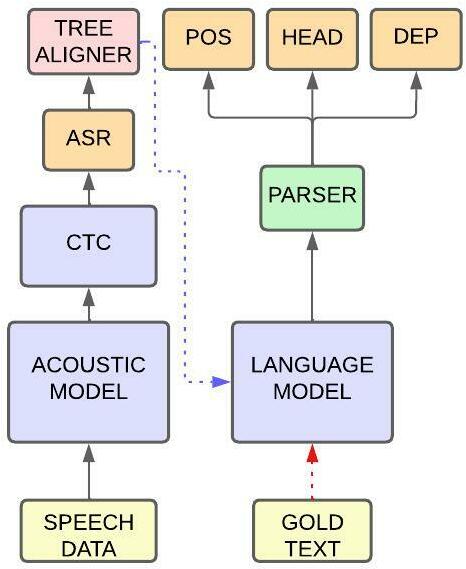}
        \caption{The two baseline models based on a pretrained language model, blue arrow is \textcolor{blue}{\PIPELINE{}} (predicted transcription), read arrow is \textcolor{red}{\TEXT{}} (gold transcriptions).}
        \label{fig:parser2}
    \end{subfigure}
    \caption{Overview of architectures with the 4 settings described in Section~\ref{sec:expe}.}
    \label{fig:experimental_settings}
\end{figure}

\section{Introduction}
Dependency parsing is a central task in natural language processing (NLP).
In the NLP community, it has mostly been addressed on textual data, either natively written texts or sometimes speech transcriptions.
Yet, speech is the main form of communication between humans, as well as arguably one of the most realistic types of linguistic data, which motivates the design of NLP systems able to deal directly with speech, both for applicative purposes and to construct corpora annotated with linguistic information.
When parsing speech \textit{transcriptions}, most prior work has focused on disfluency detection and removal \citep{charniak-johnson-2001-edit,johnson-charniak-2004-tag,rasooli-tetreault-2013-joint,honnibal-johnson-2014-joint, jamshid-lou-etal-2019-neural}, in an effort to `normalize' the transcriptions and make them suitable input for NLP systems trained on written language.
Using only transcriptions as input is a natural choice from an NLP perspective: it makes it possible to use off-the-shelf NLP parsers `as is'.
However, predicted transcriptions can be very noisy, in particular for speech from spontaneous conversations. Furthermore, transcriptions are abstractions that contain much less information than the speech signal. The prosody, and the pauses in the speech utterances are very important clues for parsing \citep{price1991use} that are completely absent from transcriptions. 
Hence, we address speech parsing using only the speech signal as input. 
With the popularization of self-supervised method and modern neural network architecture (pretrained transformers), both speech and text domains now use similar techniques \citep{chrupala-2023-putting}.
This convergence of methodology has raised interest in other applications of speech models to go beyond `simple' speech recognition.
Thus, addressing classical NLP tasks directly on speech is a natural step and design NLP tools able to deal with spontaneous speech, arguably the most realistic type of linguistic production. In short, Our contributions are the following:
\begin{itemize}[noitemsep, topsep=0pt]
    \item we introduce a graph-based end-to-end dependency parsing algorithm for speech;
    \item we evaluate the parser on Orféo, a large treebank of spoken French that features spontaneous speech, and compare its performance to pipeline systems and to a parsing-as-tagging parser;
    \item we release our code at \url{https://github.com/Pupiera/Growing_tree_on_sound}.\footnote{The code is also archived at \url{https://doi.org/10.5281/zenodo.11474162}.}
\end{itemize}

\section{Parsers and pre-trained models}
\label{sec:parser}

We define speech parsing as the task of predicting a dependency tree from an audio signal corresponding to a spoken utterance.\footnote{For the sake of simplicity, we will use the term `sentence' in the rest of the article, even though the very definition of a sentence is debatable in the spoken domain.}

Our parser is composed of 2 modules (Figure~\ref{fig:parser1}): (i) an acoustic module that is used to predict transcriptions and a segmentation of the signal in words and (ii) a parsing module that uses the segmentation to construct audio word embeddings and predict trees.

\paragraph{Word level representations from speech}
To extract representations from the raw speech, we use a pre-trained wav2vec2 model trained on seven thousand hours of French speech: \texttt{LeBenchmark7K}\footnote{\url{https://huggingface.co/LeBenchmark/wav2vec2-FR-7K-large}} \cite{PARCOLLET2024101622}.
Parsing requires word-level representations. We use the methodology of \citet{pupier22_interspeech} to construct audio word embeddings from the implicit frame level segmentation provided by the CTC speech recognition algorithm \cite{10.1145/1143844.1143891-CTC}.
The method consists in combining the frame vectors corresponding to a single predicted word with an LSTM.

\paragraph{Graph-based parsing}
We use the audio word embeddings --whose construction is described above-- as input to our implementation of a classical graph-based biaffine parser \citep{dozat2016deep}: (i) compute a score every possible arc with a biaffine classifier and (ii) find the best scoring tree with a maximum spanning tree algorithm.

\paragraph{Sequence labeling}
The sequence labeling parser follows \citet{pupier22_interspeech} and is based on the \deplabel{} approach \cite{gomez-rodriguez-etal-2020-unifying,strzyz-etal-2020-bracketing}, specifically
the \texttt{relative POS-based encoding} \cite{strzyz-etal-2019-viable}.
This method reduces the parsing problem to a sequence labeling problem. 
The head of each token is encoded in a label of the form $\pm$Integer@POS.
The integer stands for the relative position of the head considering only words of the POS category.
Eg., \texttt{-3@NOUN} means that the head of the current word is the third noun before it.

\section{Dataset}

We use the CEFC-Orféo treebank \cite{benzitoun2016projet}, a dependency-annotated French corpus composed of multiple subcorpora \citep{cfpp2000,11403/clapi/v1,11403/tcof/v2.1, OFROM, Fleuron, FON, Coralrom, delic:halshs-01388193,francard2009corpus,kawaguchi,husianycia:tel-01749085}, and released with the audio recordings.
The treebank consists of various types of interactions, all of which feature spontaneous discussions, except for the French Oral Narrative corpus (audiobooks).
Orféo features many types of speech situations (eg.\ commercial interactions, interviews, informal discussions between friends) and is the largest French spoken corpus annotated in dependency syntax.
The annotation scheme has been designed specifically for Orféo \cite{benzitoun2016projet} and differs from the Universal Dependency framework in many regards (in particular: its POS tagset is finer-grained, whereas the syntactic function tagset has only 14 relations).
The syntactic annotations of Orféo were done manually for 5\% of the corpus and automatically for the rest of the corpus.
The train/dev/test split we use makes sure that the test section only contains gold annotations.
Nevertheless, the subcorpora with gold syntactic annotations correspond to low-quality recordings, which makes them a very challenging benchmark.

\section{Experiments}
\label{sec:expe}

\begin{table*}[t]
    \centering
    \resizebox{\textwidth}{!}{
    \begin{tabular}{l r r r r r l l}
        \toprule
         Model & WER$\downarrow$ & CER$\downarrow$ & POS$\uparrow$ & UAS$\uparrow$ &LAS$\uparrow$ & Parameters & Pre-training \\
         \midrule
         \AUDIO{} \SEQ{}   & 35.9 & 22.3 & 73.0 & 65.7 & 60.4 & 315M + 34.9M & \wav \\
         \AUDIO{} \GRAPH{} & \textbf{35.6} & \textbf{22.1} & \textbf{73.1} & \textbf{66.0} & \textbf{60.9} & 315M + 34.9M & \wav \\
         \midrule
         \ORACLE{} \SEQ{}  & 36.3 & 22.2 & 75.6 & 68.7 & 62.7 & 315M + 34.9M & \wav\\ 
         \ORACLE{} \GRAPH{}& \textbf{35.6} & \textbf{22.2} & \textbf{77.4} & \textbf{73.3} & \textbf{67.5} & 315M + 34.9M &\wav\\
         \midrule
         \PIPELINE{} \SEQ{} &35.6  &22.0 &70.8 &63.8 &58.4 & 314M + 110M + 39.2M& \wavbert\\ 
         \PIPELINE{} \GRAPH{} &35.6 &22.0 &69.3 &60.5 &53.1 & 314M + 110M + 41.4M & \wavbert\\
         \PIPELINE{} \HOPS{} &35.6  &22.0  &\textbf{72.4} &\textbf{65.8} &\textbf{61.0} & 314M + 110M + 100M& \wavbert\\
         \midrule
         \TEXT{} \SEQ{} & 0 & 0&96.9 &88.8 & 85.7& 110M + 39.2M& \bert\\
         \TEXT{} \GRAPH{} & 0 & 0& 95.1 & 87.4 & 84.0 &110M + 41.4M& \bert \\
         \TEXT{} \HOPS{} & 0 & 0 & \textbf{98.2}& \textbf{90.3} & \textbf{87.7} &110M + 100M &\bert \\
         \bottomrule
    \end{tabular}
    }
    \caption{Evaluation on the full Orféo test set with the settings described in Section~\ref{sec:expe}.}
    \label{tab:comparaison}
\end{table*}

\begin{table*}[t]
    \centering
    \resizebox{\textwidth}{!}{
    \begin{tabular}{l r r r r r l l}
        \toprule
         Model & WER$\downarrow$ & CER$\downarrow$ & POS$\uparrow$ & UAS$\uparrow$ &LAS$\uparrow$ & Parameters & Pre-training  \\
        \midrule
         \AUDIO{} \SEQ{}   & 31.0 & 18.4  &\textbf{77.1}  &70.2  &65.2  & 315M + 34.9M& \wav  \\
         \AUDIO{} \GRAPH{} & \textbf{30.6} & \textbf{18.2}  &77.0  &\textbf{70.9}  &\textbf{66.2}  & 315M + 34.9M& \wav\\
         \midrule
         \ORACLE{} \SEQ{}  &\textbf{30.9} &\textbf{18.6} & 78.3 & 71.9  & 66.2 &315M + 34.9M& \wav\\ 
         \ORACLE{} \GRAPH{}&31.4 &19.2 & \textbf{79.8} & \textbf{76.0}  & \textbf{70.4} &315M + 34.9M& \wav\\
         \midrule
         \PIPELINE{} \SEQ{}  & 30.5 & 18.2 & 74.7 &67.7 &62.4 & 314M + 110M + 39.2M& \wavbert \\ 
         \PIPELINE{} \GRAPH{}& 30.5 & 18.2 & 73.5 &64.2 &57.3 & 314M + 110M + 41.4M& \wavbert\\
         \PIPELINE{} \HOPS{} & 30.5 & 18.2 & \textbf{76.3} &\textbf{69.4} &\textbf{64.6} & 314M + 110M + 100M& \wavbert \\
         \midrule
         \TEXT{} \SEQ{}   & 0 & 0 & 94.5 & 86.7 & 83.1 & 110M + 39.2M& \bert\\
         \TEXT{} \GRAPH{} & 0 & 0 & 96.8 & 88.3 & 84.5 &110M + 41.4M& \bert\\
         \TEXT{} \HOPS{}  & 0 & 0 & \textbf{98.2} & \textbf{90.3} & \textbf{87.1} &110M + 100M& \bert \\ 
         \bottomrule
    \end{tabular}
    }
    \caption{Evaluation on the Valibel corpus (a subset of the test set).}
    \label{tab:zoom_valibel}
\end{table*}

\paragraph{Experimental settings}

Our experiments aim at: (i) comparing our graph-based parser to the seq2label model,
(ii) comparing to pipeline approaches with text-based parsers, and
(iii) assessing the robustness of word representations with control experiments: using word boundaries (provided in the corpus) as input for the audio models and gold transcriptions for the text-based model. We compare the following settings (illustrated in Figure~\ref{fig:experimental_settings}):
\begin{itemize}[noitemsep]
    \item \AUDIO: Access to \textbf{raw audio} only, the model creates word-level representation from the acoustic model as described in Section~\ref{sec:parser}.
    \item \ORACLE: Access to \textbf{raw audio} and \textbf{silver}\footnote{The corpus contained word-level timestamps that have been automatically constructed through forced alignment.} \textbf{word-level timestamps}, making it easier to create word representations and mitigating the impact of the quality of the speech recognition on parsing.
    \item \PIPELINE: Access to \textbf{predicted transcriptions} from the acoustic model only, then a language model uses the transcriptions as input for parsing. 
    The training trees are modified to take into account any deletion and insertion of words. 
    However, as for the speech approach, deletion or insertion penalizes the global score of the model since the model is evaluated against the gold transcriptions and not the modified one.
    The drawback of this approach is that no information about prosody or pauses is available.
    \item \TEXT: Access to \textbf{gold transcriptions}: this unrealistic setting provides an upper bound performance in the ideal case (perfect ASR).
\end{itemize}
Both \PIPELINE{} and \TEXT{} settings use a French BERT model: \texttt{camembert-base}\footnote{\url{https://huggingface.co/almanach/camembert-base}} \cite{martin-etal-2020-camembert} to extract contextualized word embeddings.
For \PIPELINE{} and \TEXT{} settings, on top of our implementations, we use \hops{} \cite{grobol-crabbe-2021-analyse}, an external state-of-the-art graph-based parser. The \hops{} parser uses a character-bi-LSTM in addition to BERT to produce word embeddings, whereas our implementation does not (in an effort to make both versions of our parser, text-based and audio-based, as similar as possible).

\begin{table}[t]
    \resizebox{\columnwidth}{!}{
    \begin{tabular}{l|l l l l l l}
            \toprule
                        & WER$\downarrow$   & CER$\downarrow$   & POS$\uparrow$   & UAS$\uparrow$   & LAS$\uparrow$   & Parameters \\
            \midrule
             Graph-tiny  & 35.74 & 22.32 & 72.97 & 65.86 & 60.79 & 314M + 11.7M\\
             Graph-base  & 35.63 & 22.10 & 73.13 & 66.05 & 60.90 & 314M + 34.9M\\
             Graph-large & 35.60 & 22.02 & 73.17 & 65.96 & 60.67 & 314M + 67.6M\\
             \bottomrule
    \end{tabular}
    }
    \caption{Comparison of parsing metrics with the graph-based architecture and different number of parameters.}
    \label{tab:Comparaison_size_graph_based}
\end{table}
Each parsing method for each modality is trained with the same number of epochs, the same hyperparameters (see Table~\ref{tab:Hyperparam_audio} and~\ref{tab:hyperparam_text} of Appendix~\ref{sec:appendix}), and approximately the same number of parameters.
We select the best checkpoint on the development set in each setting for the final evaluation.
Our implementations use speechbrain \cite{speechbrain}.

\paragraph{Metrics}
We use classical evaluation measures: \textit{Word Error Rate} (\texttt{WER}) and \textit{Character Error Rate} (\texttt{CER}) for speech recognition, \textit{POS accuracy} (\texttt{POS}), \textit{Unlabeled Attachment Score} (\texttt{UAS}), and \textit{Labeled Attachment Score} (\texttt{LAS}) for dependency parsing.

We report results in Table~\ref{tab:comparaison} for the full corpus, and in Table~\ref{tab:zoom_valibel} for a sub-corpus of the test set (Valibel) for which speech recognition is easier.

\paragraph{Evaluation}

To evaluate our architecture, we use a modified version of the evaluation script provided by the CoNLL 2018 Shared Task.\footnote{\url{https://universaldependencies.org/conll18/evaluation.html}}
The main limitation of this evaluation protocol is that it requires the two sequences to be exactly the same, which is not the case when speech recognition is involved.
Thus, we modify this evaluation script to work even when the two sequences to evaluate are not of the same length. 
However, the modified  script requires an alignement between the 2 sequences.
For our purpose, we use an alignment based on edit distance, i.e.\ the same alignment strategy already used to compute WER.

The modified script work by following this simple set of rules, depending on the edit operations: 
\begin{itemize}[noitemsep]
    \item for word deletions: the predicted sequence is shorter, thus add a dummy token in the output sequence at the correct index to realign the sequences;
    \item for word additions: the predicted sequence is longer, thus add a dummy token in the gold sequence at the correct index to realign the sequence;
    \item for word substitutions: do nothing;
    \item The syntactic information of the inserted token must differ from that of the corresponding word in the other sequence. Thus every insertion and deletion are considered parsing errors.
\end{itemize}

\paragraph{Results: Speech recognition effect on parsing quality}
In Table~\ref{tab:comparaison}, we observe that both graph-based and seq2label-based approaches give similar results when using no additional information, which shows that the limiting factor of the model is the speech recognition, rather than the parsing. 

It is important to note that due to the nature of the speech corpus (spontaneous discussions), the \texttt{WER} is higher than what is typically expected on ASR benchmarks (usually containing `read' speech).
As a matter of fact, the ASR module used in our model reaches around 8~\texttt{WER} when trained and evaluated on CommonVoice5.1 \cite{ardila-etal-2020-common}.

Further evidence of the limitation caused by the speech recognition module is shown in Table~\ref{tab:Comparaison_size_graph_based}: changing the number of parameters of the graph-based parser does not significantly alter performance. 
Additionally, in Table~\ref{tab:zoom_valibel} we observe a clear improvement in all the parsing metrics when evaluating on a test corpus with better speech recognition performance.
The model's speech recognition ability directly affects the number of predicted tokens (some words may be deleted or added), which in turn impacts parsing.

\paragraph{Results: Difference between sequence labeling approach and graph-based approach}
It is somewhat surprising that on the text modality (\PIPELINE{}), the sequence labeling parser outperforms the graph-based approach, since this is not the case on the other modality (\AUDIO{}).
However, it does not outperform a larger graph-based model with an additional character-bi-LSTM such as \hops{}. The character bi-LSTM may mitigate the impact of out-of-vocabulary words produced by misspelling errors from the ASR.

A hypothesis about the graph-based model performance on \AUDIO{} and the \ORACLE{} settings may be that it is able to extract more relevant syntactic information from the signal due to its global decoding than simpler approaches such as sequence labeling. 

The largest gap between the two parsing approaches occur when more information about speech segmentation is given to the models (\ORACLE{}), reducing the overall influence of the speech recognition task on parsing.

\paragraph{Transcribe then parse or directly parse ?}
The \PIPELINE{} approach with \hops{} does reach a similar performance as the \AUDIO{} model with our graph-based parser.
However, \hops{} is a more complex model not fully comparable to our graph-based parser. Moreover, it has 50\% as many parameters as the model working directly on audio, requires 2 pretrained models, and is thus more expensive to train.

Lastly, Table~\ref{tab:zoom_valibel} shows that the \AUDIO{} approach outperforms the \PIPELINE{} approach when the quality of the speech recognition improves. 
This result suggests that parsing benefits from \AUDIO{} as soon as ASR reaches reasonable quality.

\section{Conclusion}
We introduced a graph-based speech parser that takes only the raw audio signal as input and assessed its performance in various settings and in several control experiments.
We show that a simple graph-based approach with wav2vec2 audio features is on a par with or outmatches a more complex pipeline approach that requires two pretrained models.

From control experiments (\ORACLE{}), we show that acquiring good quality word representations directly from speech is the main challenge for speech parsing.
We will focus future work on improving the quality of word segmentation on the speech signal.

\section*{Limitations}

We only evaluate our parsers on French, due to the availability of a large treebank, hence our conclusions should be interpreted with this restricted scope.
We plan to extend to other languages and treebanks in future work.

We did not do a full grid search for hyperparameter tuning, due to computational resource limitations and environmental considerations, although we dedicated approximately the same computation budget to each model in a dedicated setting.
However, we acknowledge that not doing a full hyperparameter search may have affected the final performance of the parsers. 

\section*{Acknowledgements}

This work is part of the PROPICTO project (French
acronym standing for PRojection du langage
Oral vers des unités PICTOgraphiques), funded by the Swiss National Science Foundation (N°197864) and the French National Research Agency (ANR-20-CE93-0005). 
MC gratefully acknowledges the support of the French National Research Agency (grant ANR-23-CE23-0017-01).

\bibliography{custom}

\appendix

\section{Training Details}
\label{sec:appendix}

Table~\ref{tab:Hyperparam_audio} and \ref{tab:hyperparam_text} describe in more detail the hyperparameters used for each parser for the different sets of modalities.

\begin{table*}[th]
    \centering
    \begin{tabular}{c|c c}
         Parser & SEQ & \GRAPH{} \\
         \hline
          Epoch & 30      & 30          \\
          Batch size & 8  & 8            \\
          \hline
          \multicolumn{3}{c}{Tuning parameters }\\
          \hline
          Learning rate & 0.0001 & 0.0001  \\
          Optimizer &AdaDelta &AdaDelta \\
          Model name & \multicolumn{2}{c}{LeBenchmark7K} \\
          \hline
          \multicolumn{3}{c}{Encoder }\\
          \hline
          Encoder layer & 3 &3  \\
          Dropout &0.15 & 0.15\\ 
          Encoder Dim & 1024  & 1024\\
          Activation & LeakyReLU  & LeakyRelu\\
          \hline
          \multicolumn{3}{c}{Fusion LSTM } \\
          \hline
          Layer & 2 &2 \\
          Dim & 500 & 500\\
          Bidirectional & False & False  \\
          Bias & True  & True\\ 
          \hline
          \multicolumn{3}{c}{LSTM parser} \\
          \hline
          Layer & 2  & 3\\
          Dim & 800 & 768\\
          Bidirectional & True & True \\
          \hline
          \multicolumn{3}{c}{Labeler (\SEQ{})}\\
          \hline
          Dim & 1600 & \\
          Layer & 1 &\\
          Linear head dim arc & 846 &\\
          Linear head dim POS & 23 &\\
          Linear head dim label & 19 &\\
          \hline
          \multicolumn{3}{c}{Arc MLP (\GRAPH{})}\\
          \hline
          Dim && 768\\
          Layer && 1\\
          Linear head dim && 768 \\
          \hline
          \multicolumn{3}{c}{Label MLP (\GRAPH{})}\\
          \hline
          Dim && 768\\
          Layer && 1\\
          Head dim && 768\\
          \hline
          \multicolumn{3}{c}{POS MLP (\GRAPH{})}\\
          \hline
          Dim && 768\\
          Linear head dim && 24 \\ 
    \end{tabular}
    \caption{\AUDIO{} and \ORACLE{} \SEQ{} and \GRAPH{} hyperparameters.}
    \label{tab:Hyperparam_audio}
\end{table*}

\begin{table*}[t]
    \centering
    \begin{tabular}{c|c c c}
         Parser& \SEQ{} & \GRAPH{} & \HOPS{}  \\
         \hline
          Epoch & 40      & 40   & 40       \\
          Batch size & 32  & 32    & 32      \\
          \hline
          \multicolumn{4}{c}{Tuning parameters }\\
          \hline
          Learning rate & 0.001 & 0.001 & 0.00003 \\
          optimizer &Adam &Adam &Adam \\
          Embedding & Last layer &Last layer & Mean First 12 layers \\
          Embedding dim &768 &768 &768 \\
          BERT & \multicolumn{3}{c}{camembert\_base} \\
          \hline
          \multicolumn{4}{c}{Char Bi-LSTM \HOPS{}} \\
          \hline
          Embedding dim & & & 128\\
          \hline
          \multicolumn{4}{c}{Word Embedding \HOPS{}} \\
          \hline
          Embedding dim & & & 256\\
          \hline
          \multicolumn{4}{c}{LSTM parser} \\
          \hline
          Dim &768 &768 & 512 \\
          Layers &3 & 2 & 3\\
          Bidirectional &True & True &True \\
          \hline
          \multicolumn{4}{c}{Labeler (\SEQ{})}\\
          \hline
          Dim & 1536 & & \\
          Layer & 1 & &\\
          Linear head dim arc & 846 & &\\
          Linear head dim POS & 23 & &\\
          Linear head dim label & 19 & &\\
          \hline
          \multicolumn{4}{c}{Arc MLP (\GRAPH{} and \HOPS{})}\\
          \hline
          Dim && 768& 1024\\
          Layer && 1&2\\
          Linear head dim && 768& 768\\
          \hline
          \multicolumn{4}{c}{Label MLP (\GRAPH{})}\\
          \hline
          Dim && 768& 1024\\
          Layer && 1&2\\
          Head dim && 768& 768\\
          \hline
          \multicolumn{4}{c}{POS MLP (\GRAPH{})}\\
          \hline
          Dim && 768& 1024\\
          Linear head dim && 24& 24 \\ 
          
    \end{tabular}
    \caption{\PIPELINE{} and \TEXT{} \SEQ{}, \GRAPH{} and \PIPELINE{} hyperparameters.}
    \label{tab:hyperparam_text}
\end{table*}

\end{document}